# Multi-camera Multi-object Tracking

Wenqian Liu
Northeastern University
liu.wenqi@husky.neu.edu

Octavia Camps
Northeastern University
camps@coe.neu.edu

Mario Sznaier
Northeastern University
msznaier@coe.neu.edu

## Abstract

*In this paper, we propose a pipeline for multi-target visual tracking under multi-camera system. For multi-camera system tracking problem, efficient data association across cameras, and at the same time, across frames becomes more important than single-camera system tracking. However, most of the multi-camera tracking algorithms emphasis on single camera across frame data association. Thus in our work, we model our tracking problem as a global graph, and adopt Generalized Maximum Multi Clique optimization problem as our core algorithm to take both across frame and across camera data correlation into account all together. Furthermore, in order to compute good similarity scores as the input of our graph model, we extract both appearance and dynamic motion similarities. For appearance feature, Local Maximal Occurrence Representation(LOMO) feature extraction algorithm for ReID is conducted. When it comes to capturing the dynamic information, we build Hankel matrix for each tracklet of target and apply rank estimation with Iterative Hankel Total Least Squares(IHTLS) algorithm to it. We evaluate our tracker on the challenging Terrace Sequences from EPFL CVLAB as well as recently published Duke MTMC dataset.*

## 1. Introduction

Stated back to 2009, after severe terror attack in New York, people decoded to produce more efficient way to detect terrorist. With the introduction of surveillance cameras into daily life, polices are able to monitor society security by looking at a computer screen. Researches aim to propose useful algorithms and tools to support human detecting suspicious target, recognize required target, and even tracking the target. Nowadays, as high quality high frame rate surveillance cameras being widely used, much more efficient methods that yield higher accuracy are needed. Within a decade, plenty of well defined detectors and trackers with competitive performance are proposed. However, most of them are focusing on single camera scenario. Another popular topic for multi-camera scenario is person re-identification. By looking at different detections of the same person maybe from different camera, reID algorithms need to extract representative features of the target and recognize it whenever the same target appears.

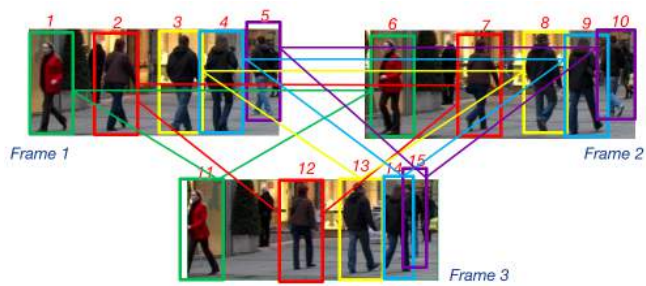

Figure 1. Example of how we forming our tracking problem as glabal maximum clique problem. Given the boundingboxes of each target of each frame, our problem is to find cliques that stitch the same target from different frames(from the same camera or different camera) based on their appearance and motion similarities. For example, the three green boundingboxes detected for as a lady in red walking towards left on the very left hand side in the frames, are picked up and stitch together as a final tracklet of that lady.

In our paper, we are cracking a multi-camera scenario multi-target tracking problem by adopting relative algorithms from reID as well as useful control system tools. We aim to solve this problem in a offline first, and later on if possible, extend it into real time tracking system. When given boundingboxes for all the targets within one video sequence, our algorithm forms a maximum clique problem based on graph theory to take all information from input into consideration. A mixed-binary linear optimization program is chosen in computing tracking result. We test our algorithm on two datasets: EPFL Terrace Sequence and Duke MTMC. The reason we choose these two datasets is because the former one represents scenario when overlap exists across cameras, while the latter represents when no overlap or only a little overlap exists. As shown in figure 1, the example is explaining clearer how we relate the global maximum cliques problem to a multi-camera multi-target tracking problem.

The rest of the paper will be presented as follows. In the second section, a related literature research will be introduced. Followed by the third section that we will mainly focus on showing our proposed framework. Later in the fourth section, we will show some experiments on our proposed algorithm and discuss a little bit of the result. In the last section, we will conclude the paper and show some possible future works.

## 2. Related Work

As multi-camera system tracking problems becoming more and more popular, new algorithms are generated and new multi-camera datasets proposed. Even new ways of evaluating multi-camera trackers' performance are interesting topics. There are mainly two types of approaches for multi-camera system tracking. The first one is to do information association inter-camera and then across camera. The second one is to globally consider all input detections. Which is the approach that this paper adopts.

There are a few papers that are working on global approach for multi-camera system multi-object tracking. A general way of forming global tracking problem is to regard all input detections as a graph. The edge weights between nodes(detections) in the graph is based on how similar the detections are. In order to compute accurate similarities, superior feature extraction algorithm is required to capture most representative features from detections. In paper [3], the authors adopted re-ID feature extraction method for edge weight and then applied min cut/max flow algorithm for tracking. Another group of researcher from UCF published [4]. In their paper, they presented a global maximum clique optimization algorithm(GMMCP). They compute the edge weights based on both appearance similarity given by comparing histogram and motion similarity given by constant velocity. This paper is proposed based on one of their previous works [10] that proposed the GMCP algorithm. The main difference between the two algorithm is that GMMCP compute the cost function for multiple cliques of tracklets at one shot. Interestingly, [9] is published in a similar manner by researchers from Duke University. Although the global fashion for information association they use is the same as [4], they only use detection's appearance feature for edge weights computation. Since it is intuitively to combine appearance similarity and motion similarity, in our paper hankel motion IHTLS [5] and re-ID LOMO feature [7] is exploited combined with GMMCP.

Multi-camera system tracking problem still remains as very new topic comparing to classical single-camera tracking. As a result, new datasets and evaluation metrics aim at multi-camera scenario are evolving. What's more, multi-camera datasets can capture mainly two types of scenarios. One is that multiple cameras look at the same scene. In other words, cameras are fully overlapping with each other. A representative dataset of this kind is video sequences produced by EPFL CVLAB [2] [6]. The second type of dataset is cameras has very little or even non overlap between each other. One newly proposed dataset is called Duke MTMC [8] [11]. Within their paper, they proposed both a new dataset and a new way for multi-camera tracking evaluation.

## 3. Proposed Framework

In this paper, we adopt a tracking by detection fashion for solving a two-camera system multi-target tracking problem. We start with bounding boxes in each frame given by state-of-art detector. Then form them into short tracklets of each targets within non-overlapping small segments of video, we denote these tracklets as low-level tracklets. Each tracklet has length of 7 to 10 frames long. After this, every few of these small temporal segments are grouped into clusters as picking by a sliding window manner. The sliding window size we choose is 5, and there are a 3-cluster overlap between every two sliding windows. All the low-level tracklets within the chosen cluster will become the input of the generalized maximum multi clique optimization problem(GMMCP). This algorithm is finding the maximum possible cliques within the graph based on edge weights between every two low-level tracklets. The edge weights are given by computing a similarity score between the two tracklets. Both appearance similarity and motion similarity are obtained by adopting Local Maximal Occurrence Representation(LOMO) as feature extraction algorithm and Iterative Hankel Total Least Squares(IHTLS) as motion extraction algorithm. Thus the final output from GMMCP will be a much longer tracking result across frames within one cluster, and at the same time hopefully across the whole video.

As follows, we will discuss in detail of the algorithms used by our proposed pipeline.

### 3.1. LOMO

Local Maximal Occurrence Representation [7] is a useful and fairly new appearance feature extraction algorithm specifically proposed in 2015 in Person Re-identification field of study. Given a detection image, by analyzing horizontally the occurrence of different local features from small patches, the LOMO feature tries to make one stable representation for each detection in order to maximize the occurrence against viewpoint changes.

In our pipeline, we input our detections one by one. Then each detection will be separate into multiple bands horizontally to compute local features. And in the end, only one feature vector is generated by LOMO for each detection. The procedure is explained in figure 2.

After obtaining LOMO feature vector for each detection, we will compute the similarity between every two feature vectors of two detections and a score will be assigned.

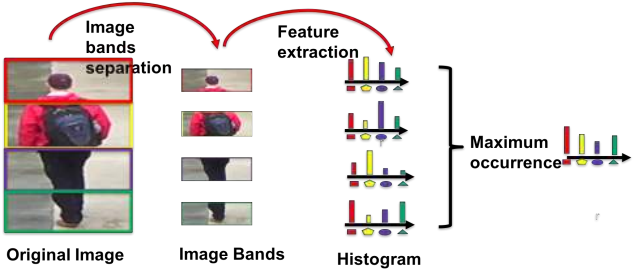

Figure 2. Each input bounding box is separated into 4 strips, and each strip is used to generate one LOMO feature. The final feature is the concatenation of the four strips.

### 3.2. IHTLS

Iterative Hankel Total Least Squares algorithm is proposed based on Hankel Total Least Squares(HTLS) algorithm [5]. There are mainly two differences between these two algorithms. The first modification that IHTLS makes is that a binary vector is introduced as an 'indicator' of missing data. If there are any missing data occurs in the middle of a tracklet, or to say that a gap occurs within the tracklet, due to occlusion or bad detection, this 'indicator' vector will be put to fill in the gap and allow IHTLS perform inpainting to recover the missing data automatically. The second modification is by increasing the estimated rank gradually, the algorithm is ran iteratively to find the optimal rank value for the given tracklet.

Given two tracklets, fist compute their rank respectively. Then combine the two tracklet by adding a 'indicator' vector if there exits a gap between them and estimate the new longer tracklet's rank. As shown in HTLS, if the three ranks computed are the same, then these two original tracklets should belong to a same and longer trajectory with same motion. If the three ranks are not the same, then they do not belong to a same motion trajectory.

Thus in our paper, we adopt this algorithm to compare the ranks of every two low-level tracklets as the input. A similarity score is assigned by IHTLS for every two tracklets.

### 3.3. GMMCP

In order to form the Generalized maximum Multi Clique Problem, we see our tracker as a undirected graph. The nodes inside the graph represent low-level tracklets. An edge between two nodes represents the two low-level tracklets belong to the same person. The edge weight is using the similarity score computed with LOMO and IHTLS.

Now, we would like to introduce some denotations before going into the formation of GMMCP.

- Camera - k. The total number of cameras is $K$. Each camera is denoted as $k$.

- Cluster - j. The total number of clusters is $J$. Each cluster is denoted as $j$.

- Node - i. The total number of nodes is $I$. Each node is denoted as $i$.

- Dummy node: $d_{jk}$ denotes the dummy node in cluster $j$ of camera $k$.

- Edge: $e_{ijk}^{i'j'k'}$ denotes the edge between node $i$ in cluster $j$ of camera $k$ and node $i'$ in cluster $j'$ of camera $k'$ ($k$ and $k'$ are not necessarily different cameras).

Then, our GMMCP can be formed as a Mixed-Binary Integer Programming in form of:

$$\begin{aligned} \text{maximize} \quad & C^T x \\ \text{subject to} \quad & Ax = b \quad and \quad Mx \ll n \end{aligned} \quad (1)$$

Where matrix $C$ stores the edge weights, and $x$ is a mixed-binary column vector with boolean elements response to regular nodes and integer dummy nodes. To be more specific according to the formulation of GMMCP problem, we can expand the cost function into four terms:

$$\sum_{1}^{RealEdges} \overbrace{c_1 x_1} + \sum_{2}^{DummyEdges} \overbrace{c_2 x_2} + \sum_{3}^{RealNodes} \overbrace{c_3 x_3} + \sum_{4}^{DummyNodes} \overbrace{c_4 x_4} \quad (2)$$

Having th object function for our problem, now we need to define the constraints in order to make sure the solution is valid.

- Constraint 1 ensures that every three nodes picked up from three different clusters will form a clique.

$$e_{ijk}^{i'j'k'} + e_{i'j'k'}^{i''j''k''} \ll e_{ijk}^{i''j''k''} + 1 \quad (3)$$

- Constraint 2 enforces the total number of outgoing edges from one node in cluster $i$ will enter another cluster $j$ only once or zero time.

$$\sum_{i'=1}^{I^*} e_{ijk}^{i'j'k'} \ll 1 \quad (4)$$

- Constraint 3 guarantees that given H clusters in total, then N nodes, including dummy nodes, from each clusters should be selected.

$$\sum_{i=1}^{K} \sum_{j=1}^{J} \sum_{k=1}^{K} e_{ijk}^{i'j'k'} + d_{ij} = (H-1) \times N \quad (5)$$

## 4. Experiments

In order to test our proposed approach, we found two datasets that fit the best of our multi-camera multi-object problem. One is EPFL terrace sequences, the other is the very new Duke MTMC dataset. The objects for tracking are both human targets in these two datasets. The initial detection bounding boxes are all given by doppia toolbox which implements the proposed detector of paper [1] offline.

Moreover, after computing the appearance similarity and motion similarity respectively, we perform a weighted sum over the two similarity score to give the final edge weights and store in cost matrix $C$ as introduced above. We pick different appearance weight given different dataset, and the motion weight is equal to $1 - appearance_{weight}$. Also, the dummy node weight is also picked respect to $appearance_{weight}$.

### 4.1. Dataset and Evaluation Method

**EPFL Terrace Sequence** The first dataset that we tested on is from CVLAB of EPFL. The sequences were shot outside a building on a terrace. Up to 7 people evolve in front of 4 DV cameras, for around 3 1/2 minutes. The frame rate is 25 fps. The 4 cameras capture fully overlapped area of the terrace. The ground truth is given every 25 frame. By using the Tsai camera calibration also provided on the website, we can compute a 3D world coordinates for evaluation.

For multi-camera scenario, we tested our method with terrace sequence 1 camera 0 and camera 1. We adopt the standard clearMOT evaluation, which will provide MOTA and MOTP score.

**Duke MTMC Dataset and Evaluation Method** DukeMTMC is a new and large dataset mainly for multi-target multi-camera tracking problems which was first proposed in 2016 ECCV. It provides a new large scale 1080p video data set recorded by 8 synchronized cameras under 60 fps for almost 85 minutes. There are more than 7,000 single camera trajectories and over 2,000 unique identities. More than 2,000,000 manually annotated frames are provided within a certain region of interest as groundtruth. Within these 8 cameras, only camera 2 and camera 8 share a small portion of overlapping scene. The rest 6 cameras are watching at different part of Duke campus and a top view is provided on their website. A more clear view of the camera topology can be seen in figure3 In addition to this huge dataset, the group of researchers also proposed a new performance evaluation method that focusing on measuring how often a system is correct about who is where, regardless of how often a target is lost and reacquired. The reason why they propose this new evaluation method, claimed by its authors, is due to the fact that the widely used standard clearMOT method fails to handle and generalize scenarios under multi-camera systems and hence yield reliable and meaningful evaluation scores.

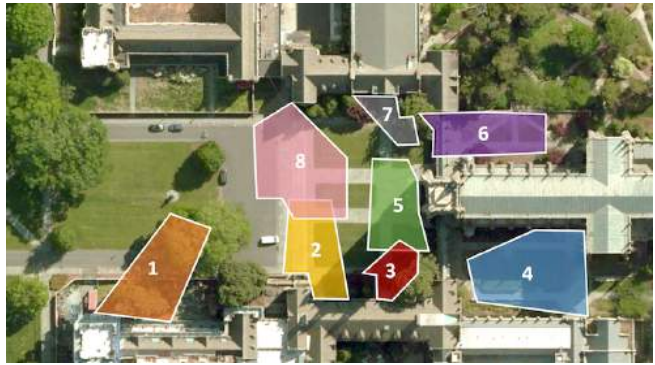

Figure 3. The camera topology provided from the website of the dataset. The different eight portions are the campus areas that are watched by different cameras receptively. As shown in the figure, the eight cameras merely share any overlapping.

Table 1. Evaluation Result on EPFL Terrace Sequence 1

|      | rateTP | rateFP | rateFN | IDswitch | MOTA |
|------|--------|--------|--------|----------|------|
| Ours | 0.42   | 0.003  | 0.53   | 60       | 0.42 |
| [12] | -      | -      | -      | -        | 0.7  |

Table 2. Evaluation Result on Duke MTMC

| Tracker    | IDF1 | IDP   | IDR  |
|------------|------|-------|------|
| MTMC_CDSC  | 60   | 68.3  | 53.5 |
| Lx_b       | 58   | 72.6  | 48.2 |
| BIPCC      | 56.2 | 67    | 48.4 |
| dirBIPCC   | 52.1 | 62    | 45   |
| PT_BIPCC   | 34.9 | 41.6  | 30.1 |
| Ours       | 55.5 | 78.89 | 44.6 |

### 4.2. Result

We fist tested our pipeline on EPFL Terrace Sequence 1 and compare to the state-of-art method proposed by [12]. We are using the groundtruth provided on EPFL website, while [12] hand labled their own groundgruth. For this dataset, we choose $appearance_{weight} = 0.7$ and $dummy_{weight} = 0.7$. The results are shown in table1. Then we tested and evaluated using Duke MTMC new dataset and their evaluation method. We compared our result to state-of-art scoreboard posted on motchallenge website. This time, a $dummy_{weight} = 0.6$ is chosen, and the result is shown in table2

As we can see from the result, the result yields by proposed algorithm cannot beat state-of-art result. One possible reason could due to our information merging algorithm is not good enough to associate the output from GMMCP and hence project back into each single camera for yielding evaluation score. Another possible reason may be that the mostion of the targets in the video are human. Human motion rank are similar to each other which results in our rank estimation algorithm fails to give meaningful similar-ity score to separate targets away.

We also estimate on the computational time for the whole pipeline. It takes a little bit more than 1 hour to run the whole pipeline. The most time-consuming part is building the similarity matrix, which takes up to 1 hour already(4138s). The second time-consuming part is Gurobi(289s). The third is forming tracklets every 10 frames(130s).

Some more qualitative results will be shown in figure4 and 5. For EPFL dataset, an ID number is assigned on top of each bounding box. The same ID across frame and camera will share the same color bounding boxes. The yellow box with numbers begin with a # represent the frame numbers. Thus the tracking result of the two camera at the same frame are showing horizontally. The tracking consistency is showing vertically instead. The result from DukeMTMC dataset is shown in figure5. As you may see, some of the bounding boxes are not labeled in the frames due to the region of interests(ROI) provided by the original dataset.

Furthermore, in order to understand how well dynamic information helps our tracking algorithm, we tune the parameter $appearance_{weight}$ from 0 to 1 to yield the plots in figure 6. When the $appearance_{weight} = 0$, we use only motion similarity for tracking, and we have the highest IDP, IDR, and IDF1 scores using MTMC evaluation method. When calculating the motion similarity, we compared euclidean distance with EMD distance and the plots are shown together in figure 6.

## 5. Conclusion

In this paper, we adopt a global information association manner for solving a multi-camera multi-target tracking problem. We first obtain our detection with a state-of-art detector based on deep learning. Then we observe our detections as a large graph and compute a globally maximum cliques optimization problem formed by mixed-integer linear program. We adopt re-ID LOMO feature for detection's appearance feature extraction method and hankel matrix based IHTLS algorithm for motion feature. The two features are combined to provide edge weights for the graph.

The algorithm is tested on two dataset: EPFL Terrance Sequence1 and Duke MTMC. The evaluation is given by standard clearMOT metric and Duke MTMC metric respectively.

As shown in the result, our algorithm still have space for improvements for tracking accuracy. The possible reasons that our tracker is not working as good as state-of-art are mainly two. First of all, a dataset with more complex motion information may be needed for our tracker to outperform others. Secondly, a better way for stitching final tracklet and merging the information got from GMMCP algorithm may be required. Thus, our future work will mainly focus on improve the two problems.

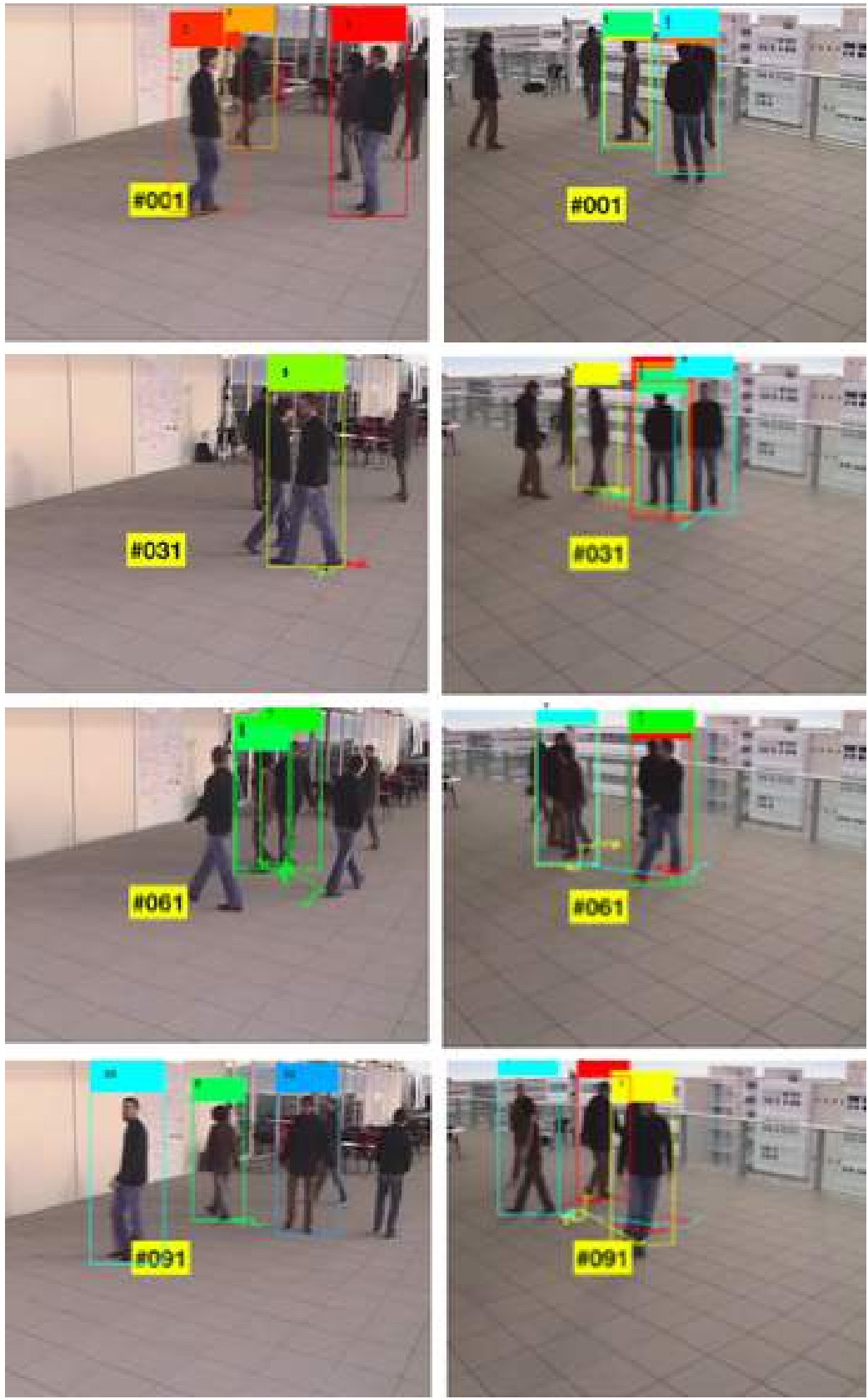

Figure 4. Qualitative result got from EPFL Terrace Sequence 1. The first column(one on the left) is from video sequence captured by camera 0, while the second column(one on the right) is from camera 1. The two cameras are watching at the same area of a terrace on a building.

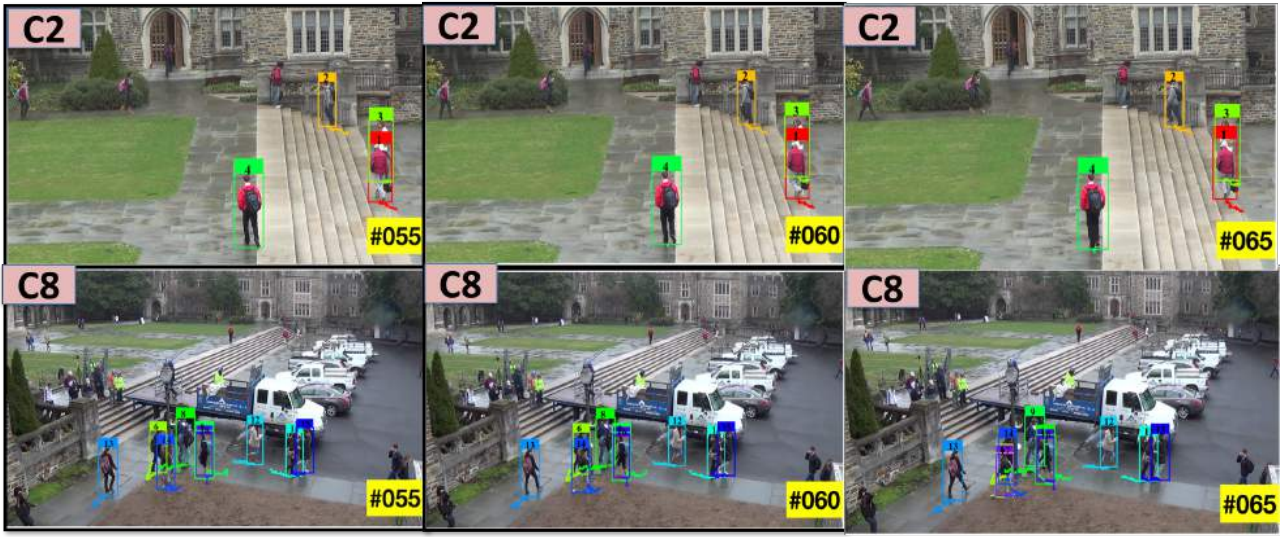

Figure 5. Qualitative result got from Duke MTMC dataset. The first row are frames from video sequence captured by camera 2, while the second row camera 8. Each column represents the same of the two cameras. As shown in the pictures, our algorithm correctly detects and tracks multiple targets.

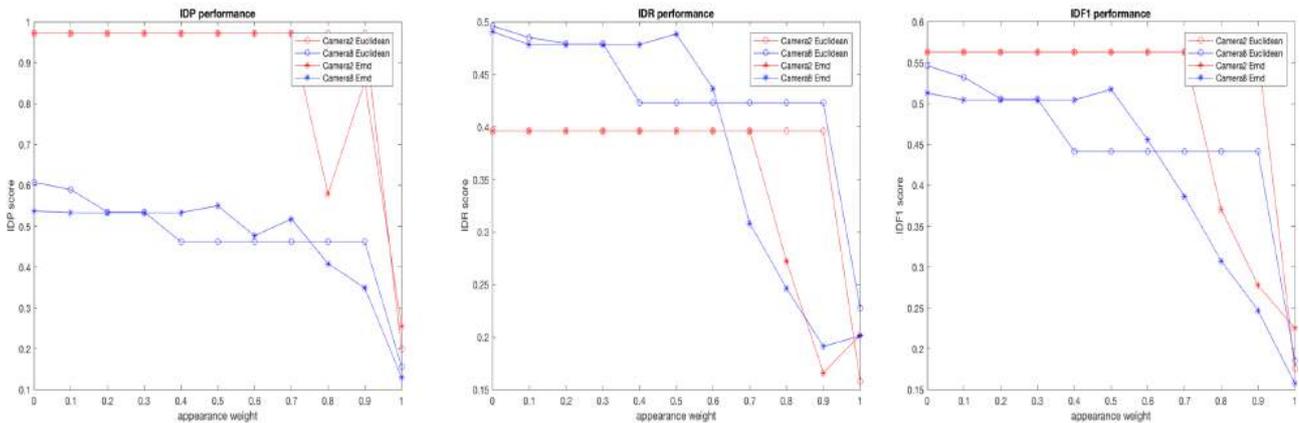

Figure 6. The first plot on the left shows the larger the appearance weight, the less motion information is involved, the smaller the IDP score. The plot in the middle shows a similar situation for IDR score, while the plot on the right for IDF1 score.